\documentclass[journal]{IEEEtran}
\usepackage{datetime}
\usepackage{times}
\usepackage{epsfig,graphicx}  
\usepackage{graphicx, amssymb}
\usepackage{amsfonts, enumerate}
\usepackage{latexsym,subfigure,makeidx}
\usepackage{url}
\usepackage{tabls}
\usepackage{algorithm}
\usepackage{algpseudocode}
\usepackage{lscape}
\usepackage[sort,space]{cite}
\usepackage{pifont}
\usepackage{tabularx}
\usepackage{rotating}
\usepackage{hhline}
\usepackage{textcomp,booktabs}
\usepackage{xcolor}
\usepackage{array}
\usepackage{amsmath,bm}
\usepackage{caption}
\usepackage{multirow}
\usepackage{verbatim}
\usepackage{amsthm}
\usepackage{framed}

\newtheorem{definition}{Definition}

\ifCLASSINFOpdf
\else
\fi
\ifCLASSOPTIONcompsoc
  \usepackage[caption=false,font=normalsize,labelfont=sf,textfont=sf]{subfig}
\else
 \usepackage[caption=false,font=footnotesize]{subfig}
\fi
\begin{document}
%
\title{Deep Reinforcement Learning for Stochastic Computation Offloading in Digital Twin Networks}
%
%

%
\author{Yueyue~Dai, \IEEEmembership{Member,~IEEE}, 
         Ke~Zhang,
         Sabita~Maharjan, \IEEEmembership{Senior~Member,~IEEE}, \\
        and~Yan~Zhang, \IEEEmembership{Fellow,~IEEE}
\thanks{This research was supported in part by the National Natural Science Foundation of China under Grant No. 61941102 and in part by the Xi’an Key Laboratory of Mobile Edge Computing and Security, under Grant No. 201805052ZD3CG36. (Corresponding author: Yan Zhang) }
\thanks{Y. Dai and K. Zhang are  with the School of Information and Communication Engineering, University of Electronic Science and Technology of China, Chengdu 611731, China (email:yueyuedai@ieee.org; zhangke@uestc.edu.cn).}
\thanks{S. Maharjan and Y.  Zhang are with Department of Informatics, University of Oslo, Norway, and also with Simula Metropolitan Center for Digital Engineering, Norway. (email: sabita@ifi.uio.no, yanzhang@ieee.org).}
}
\maketitle

	\begin{abstract}
	The rapid development of Industrial Internet of Things (IIoT) requires industrial production towards digitalization to improve network efficiency. Digital Twin is a promising technology to empower the digital transformation of IIoT by creating virtual models of physical objects. However, the provision of network efficiency in IIoT is very challenging due to resource-constrained devices,  stochastic tasks, and resources heterogeneity. Distributed resources in IIoT networks can be efficiently exploited through computation offloading to reduce energy consumption while enhancing data processing efficiency.   In this paper, we first propose a new paradigm Digital Twin Networks (DTN) to build network topology and the stochastic task arrival model in IIoT systems. Then, we formulate the stochastic computation offloading and resource allocation problem to minimize the long-term energy efficiency. As the formulated problem is a stochastic programming problem, we leverage Lyapunov optimization technique to transform the original problem into a deterministic per-time slot problem. Finally, we present Asynchronous Actor-Critic (AAC) algorithm to find the optimal stochastic computation offloading policy. Illustrative results demonstrate that our proposed scheme is able to significantly outperforms the benchmarks.

	\end{abstract}

	\IEEEpeerreviewmaketitle
\section{Introduction}

The Industrial Internet of Things (IIoT) is an enabling technology of Cyber-Physical Systems (CPSs)  that can equip the industrial units, such as  sensors, instruments, and devices with the ability to communicate and
interact with each other. The IIoT has undergone rapid technological development in recent years.
According to the report from International Data Corporation (IDC) \cite{idc2019growth},  the number of connected devices will reach  $ 41.6$ billion and these devices are predicted to generate nearly 80 zettabytes of data by 2025.  The high spread of IIoT requires industrial production towards network and digitalization.



\begin{figure}[!tp]
 	\centering
 	\includegraphics[width=3.2 in]{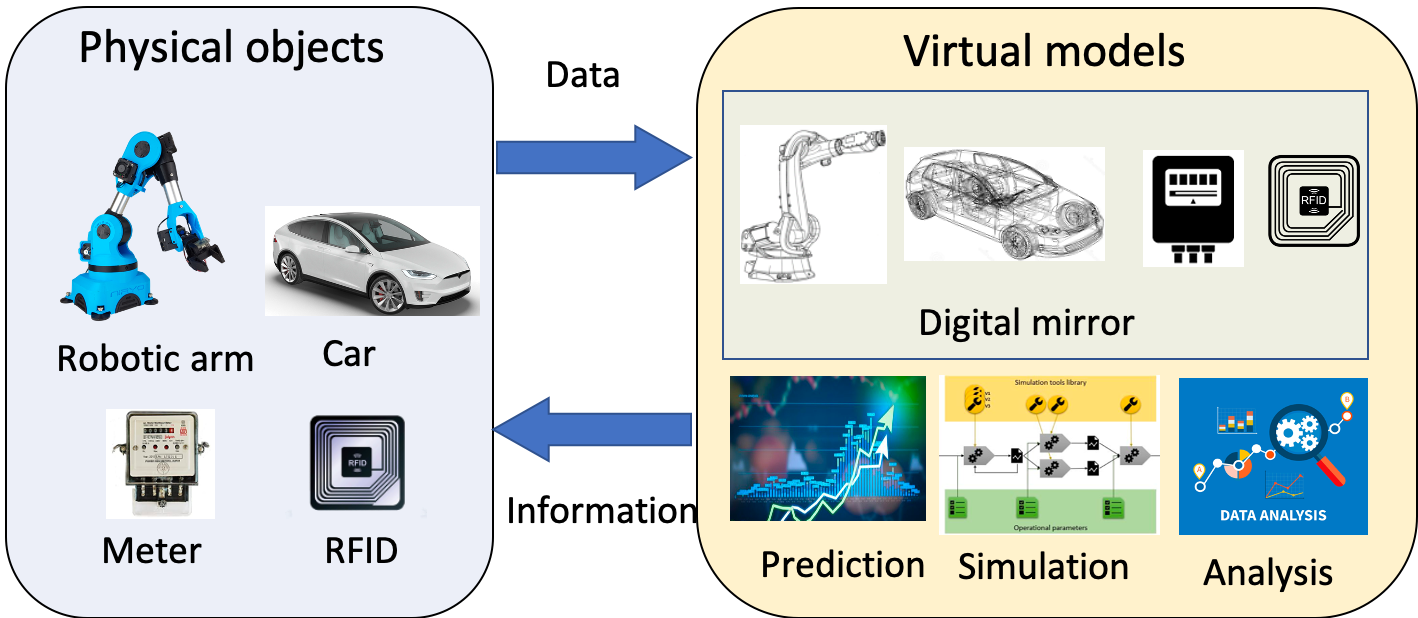}
 	\caption{Concept of digital twin}
 	\label{figure_dt}
 \end{figure}

Digital twin is a powerful technology to enable the digital transformation by creating virtual models of physical objects in the digital way, as shown in Fig.  \ref{figure_dt}.  The virtual models can understand the state of the physical entities through sensing data, so as to predict, estimate, and analyse the dynamic changes. The concept of digital twin is first proposed in \cite{glaessgen2012digital} and applied by NASA  to comprehensive diagnosis and maintenance of flight systems. Recently, digital twin has been expanded to smart cities, manufacturing  and IIoT.  Exploiting digital twin, the network topology and physical elements in IIoT can be well mirrored and we can make system management based on the  mirrored models. However, there are many technical challenges in applying digital twins to IIoT. First, massive data collected from various IIoT devices needs to be processed  timely. But the limited computing resources available at the local servers cannot support fast data processing and digital twin modelling in IIoT networks \cite{zhang2018mobile}, \cite{9145588}. 
Second, the interaction between the virtual models and the physical objects in a digital twin-enabled network requires frequent communication between them.  Moreover, since the communication is wireless, the stochastic associated with the wireless channel may result in a poor transmission link, correspondingly longer service delay.
 
  \begin{figure*}[!htp]
  	\centering
  	\includegraphics[width=6.5 in]{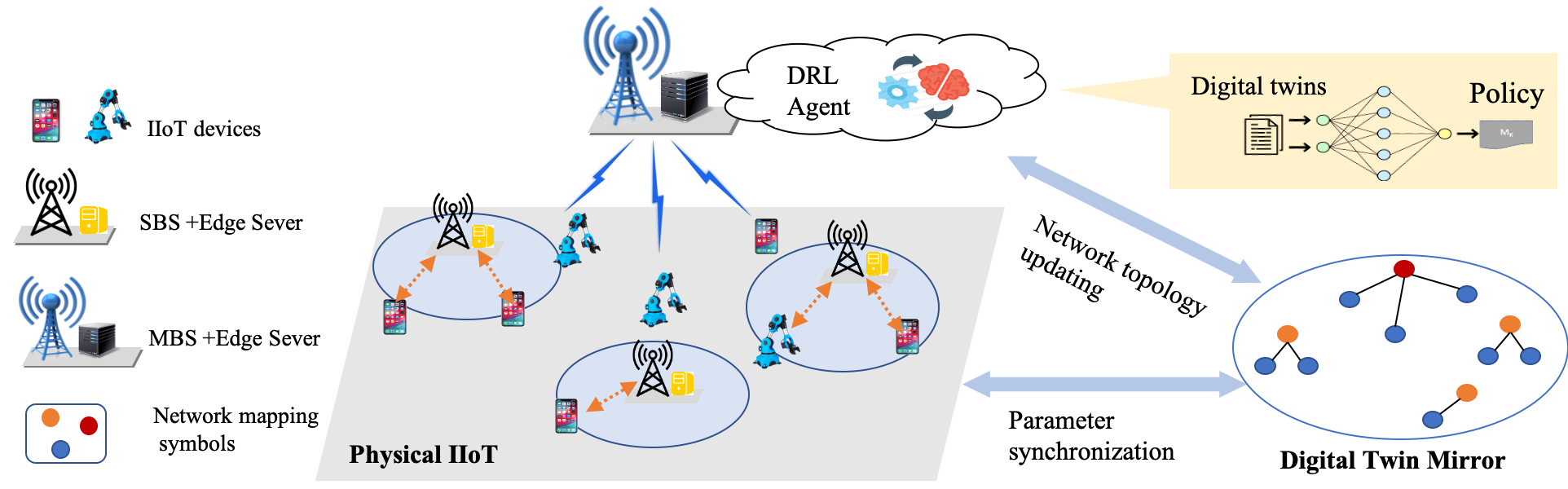}
  	\caption{The digital twin network}
  	\label{overview}
  \end{figure*}

The IIoT applications, such as data analytics and smart manufacturing, involve plenty of computation tasks.  To improve data/task processing efficiency and prolong battery lifetime of IIoT devices, computation offloading is a promising approach which offloads the collected data and computation tasks to  distributed servers to process, such as base stations, access points, and road-side units in an IIoT network \cite{7931566}, \cite{8643392}. There has been considerable amount of work focusing on computation offloading in wireless networks and vehicular networks.  The authors  in \cite{dai2018jointvel}  proposed to offload computation tasks  to  lightweight and distributed  road-side units to minimize task processing latency in vehicular networks. 
The authors in \cite{ning2019joint} proposed   a joint computation offloading, power allocation, and channel assignment problem to maximize the achievable sum rate for 5G-enabled traffic management systems. The authors in \cite{dai2018joint} proposed to offload computation tasks to multiple  distributed Small-cell Base Stations (SBS) and  Macro-cell Base Stations (MBS)  to minimize energy consumption in 5G networks.   The  above works typically assumed that each device executes a single computation task  without considering  the randomness of task arrivals \cite{mao2017stochastic},\cite{mao2019energy}.  Such assumptions make the computation offloading design not practical for IIoT networks. Since devices in IIoT networks continuously generate data, stochastic task arrival model is more reasonable and long-term computation performance needs to be considered.  Moreover, in an IIoT network with heterogeneous resources, it is challenging to jointly optimize offloading decision, transmission power, bandwidth and computation resource   while also incorporating  time-varying channel condition.

Deep Reinforcement Learning (DRL) is an emerging technique  to address problems with characterized with time-varying feature \cite{ahmed2019deep}, \cite{8998330}.   State-of-the-art works  have utilized DRL for optimizing computation offloading in wireless networks and vehicular networks. For instance, the authors in \cite{zhang2019deep} proposed  a Deep-Q Network (DQN)  based task offloading scheme to select the optimal edge  server  and the optimal transmission mode to maximize task offloading utility in vehicular networks. The authors in \cite{xie2019backscatter} proposed the double DQN based  backscatter-aided hybrid data offloading scheme to  reduce  power consumption for data transmission.  The authors in \cite{9158401} proposed a  Deep Deterministic Policy Gradient (DDPG) based computation offloading and resource allocation scheme to minimize system energy consumption in wireless networks.  These works however mainly focus on to choosing whether to execute tasks locally or offload to edge servers with a  deterministic task arrival model. These solutions are not directly applicable to IIoT networks since the task arrival model is stochastic.

 In this paper, we first propose a  Digital Twin Network (DTN) which utilizes digital twin to establish an efficient mapping between IIoT and digital systems.  In the proposed DTN network, virtual models of IIoT entities can be created by monitoring real-time states  of devices and base stations.  Then, we formulate stochastic computation offloading  problem as an optimization problem. Based on the virtual models and monitored information of digital twin, we design a DRL-based  algorithm  to solve the formulated problem.  Our main contributions in this paper are summarized as follows:
\begin{itemize}
\item  We propose  an architecture to integrate digital twin with the IIoT network to  model  network topology,  physical devices and base stations.

\item We formulate stochastic computation offloading problem as an optimization problem, and utilize  Lyapunov optimization technique to equivalently transform the original problem to a deterministic per-time slot problem.

\item We  adopt Asynchronous Actor-Critic (AAC),  a DRL-based algorithm, to solve the computation offloading  and resource allocation problem. Numerical results  demonstrate that our proposed  algorithm  significantly outperforms the benchmark policies. 
\end{itemize}

\section{System Model described by Digital Twin}

\label{sm}

\subsection{ The  Digital Twin Network}
Our digital twin network architecture is  shown in Fig. \ref{overview}, which  consists of  physical IIoT network and  digital twin. 

The physical IIoT network has three major components, i.e.,  distributed IIoT devices,  SBSs, and centralized  MBS.  Each device  collects the data from sensors and on-device applications, and they need to analyse the collected data. As data analysis is computation-intensive, devices with  limited computation capability and battery, may not be able to complete them  timely. So they have to offload these tasks to edge servers for a high level of quality of computation experience.  SBSs  are equipped with edge servers and they can provide devices computation resources. However, since an SBS often serves several devices,  to ensure the requirements of all devices are satisfied,  SBSs need to optimize computation resources, bandwidth and transmission power. 
{The MBS is  equipped with an edge server  and an DRL agent, thus the MBS has sufficient communication, computation and AI-enabled processing capabilities.  }

 The digital twin is a virtual model of physical elements and a digital representation of the physical system. Different from a virtual prototype, digital twin not only mirrors the characteristic of physical elements/system but also makes prediction, simulates the system, and can play a crucial role in optimizing the resources.    In our digital twin network , we can utilize digital twin to  (1)  construct the network topology of physical IIoT, (2) monitor network parameters and models, i.e., dynamic changes of resources and  stochastic task arrival processes,  (3) optimize offloading and resource allocation policy. Specifically, we deploy different functions on devices, SBSs and the MBS to build digital twin network.  Devices run two DT functions: data collection and parameter synchronization.  SBSs also run two DT functions: building  local virtual models of devices and SBSs, and synchronizing parameters. The main functions of the MBS are to  construct the network topology of the physical network and  to design offloading and scheduling policy.

\subsection{ Network  Model}
\label{secDT}
Based on digital twin, the digital representation (i.e., virtual models) of the physical  network  is created (i.e., virtual world). The digital models here contain  wireless network topology, communication model between devices and base stations,  and the stochastic task queueing model.

\textit{1) Network Topology and Communication Model for DTN}

Digital twin firstly models the wireless edge network as a discrete time-slotted system. A directed graph $G = (\mathcal{U}, \mathcal{B} , \varepsilon)$  is used to represent the  network, where $\mathcal{U}=\{u_1,..,u_N\}$ and $\mathcal{B} = \{b_0,b_1,...,b_M\}$ denote the set of  devices and base stations ($b_0 $ is the index for MBS) respectively. $\varepsilon$ denotes the association between devices and base stations. That is, if device $u_i$ is connected to SBS $b_j$, the link will be recorded in edge set $\varepsilon$. 
Then, the digital twin uses a $3$-tuple $DT_i(t) $ to  characterize  devices, i.e.,
\begin{equation}
DT_i(t) = \{p_{i,max}(t),l_i(t), f_i^l\}
\end{equation}  where $p_{i,max}(t)$ denotes the maximal transmission power at time slot $t$, $l_i(t)$ denotes the current location of $u_i$, and $f_i^l$ denotes the  computation resources of local server.  Similarly, the digital twin uses a $3$-tuple $DT_j(t) $ to characterize base stations, i.e.,
\begin{equation}
DT_j(t) = \{l_j(t),w_j, f_j^e\}
\end{equation}
 where $l_j(t)$ denotes the current location of  $b_j$, $w_j$ denotes the channel  bandwidth of $b_j$,  $f_j^e$ denotes the computation resource.
 
The task offloading between devices and base stations is facilitated through wireless communication. According to \cite{dai2018joint}, wireless data rate is related to spectrum, interference and bandwidth. In wireless  networks, to utilize spectrum efficiently, SBSs  reuse the MBS's frequency resource and  Orthogonal Frequency Division Multiple Access (OFDMA)  is often adopted to suppress  the interference.  Thus,  the interference  between the MBS and SBSs can be ignored. Here, we consider devices communicate with  the nearest base station to perform computation offloading.  $\gamma_j^s$ is defined as the coverage radius of SBS $b_j$. If the distance between  device $u_i$ and SBS $b_j$ is less than $\gamma_j^s$ (i.e.,  $r_{ij}^s(t)<\gamma_j^s$), device $u_i$  can  communicate with $b_j$.
 The  wireless communication data rate  between  device $u_i$ and  SBS $b_j$ can be expressed as 
  \begin{equation}
  \label{rate_sbs}
  R_{ij}^s(t) ={w_{ij}(t)}\log(1+\frac{p_i(t)h_{ij}^s(t)r_{ij}^s(t)^{-\alpha}}{\sigma^2+I}), 
  \end{equation}
  where $w_{ij}(t)$ ($w_{ij}(t)\leq w_j$) is  the  bandwidth that SBS $b_j$ allocates to device $u_i$ at time slot $t$, $h_{ij}^s(t)$ is the current channel gain,  $\alpha$ is path loss exponent,  $\sigma^2$ is noise power, and $r_{ij}^s(t)$ is calculated based on the location of $l_i(t)$  and $l_j(t)$ (i.e., $r_{ij}^s(t) = \parallel l_i(t)-l_j(t)\parallel$). $I = \sum_{i'\in\mathcal{U}/\{i\},j'\in\mathcal{B}/\{j\}}p_{i'}(t)h_{i'j'}^s(t)r_{i'j'}^s(t)^{-\alpha} $ is the interference  from other  SBSs.

  If  device $u_i$ does not lie within the coverage of any SBS,  it will communicate  with the MBS. The  wireless communication data rate between  device $u_i$ and the MBS is
 \begin{equation}
 \label{rate_mbs}
 R_{i0}^m (t) ={w_{i0}(t)}\log(1+\frac{p_i(t)h_{i0}^m(t)r_{i0}^m(t)^{-\alpha}}{\sigma^2}), 
 \end{equation}
 where $w_{i0}(t)$ ($w_{i0}(t)\leq w_0$)  is the channel bandwidth between  device $u_i$ and the MBS at time slot $t$, $h_{i0}^m(t)$ is the channel gain  between  device $u_i$ and the MBS at time slot $t$, $r_{i0}^m(t) = \parallel l_i(t)-l_0(t)\parallel$ is the distance between  device $u_i$ and the MBS.

\textit{2) Stochastic Task Queueing Model for DTN} 

At the beginning of time slot $t$,  device $u_i$ generates and stores $\lambda_i(t)$ (bits/slot)  of computation tasks into the local dataset. Without loss of generality, we assume $\lambda_i(t)$ in different time slots are independent, and $\mathbb{E}[\lambda_i(t)] =\lambda$. Due to the limitation of computation resources, each device executes part of the computation tasks at its local server and offloads part of them to the associated base station.  The rest will be queueing in the local task buffer and we consider the buffer has sufficient capacity. We denote the size of  the computation tasks that  is executed locally as $D_i^l(t)$ and  the size of  the computation tasks offloaded to base station $b_j (j\in\mathcal{B})$ as $D_{ij}^e(t)$. The queue length of  local task buffer at the beginning of time slot $t$ on device $u_j$ is denoted as $Q_i^l(t)$ and the queue length  is dynamically updated with the following equation:
\begin{equation}
\label{queue1}
Q_i^l(t+1) = \max\{Q_i^l(t)-\Psi_i(t),0\}+\lambda_i(t)
\end{equation}
where $\Psi_i(t) = D_i^l(t)+D_{ij}^e(t)$ is the size of the computation tasks that leaves the task buffer of device $u_i$ during time slot $t$.

Each edge server also has a task buffer to store the offloaded but not yet executed task. We denote the queue length of  edge task buffer at the beginning of time slot $t$  on base station $b_j (j\in\mathcal{B})$ as $Q_j^e(t)$. 
The queue length  is dynamically updated by:
\begin{equation}
\label{queue2}
Q_j^e(t+1) = \max\{Q_j^e(t)-\Psi_j(t),0\}+\sum_{i\in\mathcal{U}}D_{ij}^e(t)
\end{equation}
where $\sum_{i\in\mathcal{U}}D_{ij}^e(t)$ is the  amount of tasks offloaded from devices during time slot $t$. $\Psi_j(t) $ is the size of the computation tasks that departs edge task buffer (i.e., executed by edge server). {According to the definition of stability in \cite{neely2010stochastic}, task queue is stable if all computation tasks satisfy  the following constraints: 
\begin{subequations}
	\begin{align}
	&\lim_{T \rightarrow \infty} \dfrac{1}{T}\sum_{t=0}^{T-1}\sum_{i \in \mathcal{U}}\mathbb{E}\{Q_i^l(t)\}<\infty \label{st1}\\
	&\lim_{T \rightarrow \infty} \dfrac{1}{T}\sum_{t=0}^{T-1}\sum_{j \in \mathcal{B}}\mathbb{E}\{Q_j^e(t)\}<\infty \label{st2}
	\end{align}
\end{subequations}	
}

\subsection{Task Offloading Model}

During time slot $t$, device $u_i$  executes  $D_i^l(t)$ locally and offloads $D_{ij}^e(t)$ to base station $b_j$.  Next, we will calculate the energy consumption of local execution and computation offloading.

 \textit{1) Local Execution}:

Let $f_i^l (t)$ denote  the computation  resource (i.e., CPU cycles per second) of device $u_i$ during  time slot $t$.  $c$ denotes the required number of CPU cycles for executing one bit of computation task.  Thus, the size of  computation tasks executed locally will be
 \begin{equation}
 \label{local_time}
 D_i^l (t)=\dfrac{\tau f_i^l(t)}{c},
 \end{equation}
 where $\tau$ is  duration of the time slot.
 
The energy consumption of unit computation resource is $\varsigma  ({f_i^{l}})^2$, where $ \varsigma $ is the effective switched capacitance depending on the chip architecture
\cite{dai2018joint}.  We denote local energy consumption for computing task $D_i^l (t)$ as  $E_i^l (t)$, which can be defined as
\begin{equation}
\label{local_energy}
E_{i}^l(t)={\varsigma  \tau f_i^{l}(t)^3.}
\end{equation}

 \textit{2) Edge Server Execution}:  
 
 Devices offload their tasks to  base stations via wireless communication. Since devices are associated with different base stations,  the offloaded tasks of device $u_i$ during time slot $t$ can be expressed as
\begin{equation}
\label{time_SBS}
D_{ij}^e(t)=
\begin{cases}
R_{ij}^s(t) \tau& j\in \mathcal{B}/\{b_0\}\\
R_{i0}^m(t) \tau&j=b_0\\
\end{cases}
\end{equation}
 
 The energy consumption in this case consists of three parts. The first one is the energy consumption of uplink wireless transmission for offloading. The second one is the computation energy consumption which is related to the allocated computation resources. The third one is the energy consumption of downlink wireless transmission for offloading computation result  to  the devices. Since the size of the result is very small, we ignore the energy consumption for downlink transmission. Thus, the energy  consumption  for executing task $D_{ij}^e(t)$ on base station $b_j$   is given by 
 \begin{equation}
 \label{energy_SBS}
 E_{ij}^e(t)=p_i(t) \tau+\dfrac{D_{ij}^e(t)*c}{f_{ij }^e(t)}*\varepsilon,
 \end{equation}
 where $f_{ij}^e(t) $ is the computation  resource  that $b_j$ allocates to device $u_i$ at time slot $t$,  $\varepsilon$ is the  energy consumption for unit computation on edge servers.
 
 The total energy consumption is consisted of the energy consumption of  task execution in the local and edge servers, as well as the transmission energy consumption for computation offloading. Therefore, the total energy consumption can be expressed as 
   \begin{equation}
   \label{total}
   E^{tol}(t)=\sum_{i\in \mathcal{U}} E_i^l(t) + \sum_{i\in \mathcal{U}} \sum_{j \in \mathcal{B}} E_{ij}^e(t),
   \end{equation}

 \section{ Problem Formulation}
 In this section, we  first formulate the stochastic computation offloading problem of DTN as an optimization problem,  and then transform the  formulated problem  based on Lyapunov optimization.

\subsection {Stochastic Computation Offloading Problem}

The objective of  stochastic computation offloading  problem is to minimize  network efficiency $\eta_{EE}$. 
$\eta_{EE}$ is defined as the ratio of long-term total energy  consumption to the corresponding long-term aggregate accomplished computation tasks, i.e.,

 \begin{equation}
 \label{ee}
 \eta_{EE} = \dfrac{\lim_{T \rightarrow \infty} \dfrac{1}{T}\sum_{t=0}^{T-1}\mathbb{E}\{E^{tol}(t)\}}{\lim_{T \rightarrow \infty} \dfrac{1}{T}\sum_{t=0}^{T-1}\sum_{i \in \mathcal{U}}\sum_{j \in \mathcal{B}}\mathbb{E}\{D_i^t(t)+D_{ij}^e(t)\}}
 \end{equation}

The system operation at time slot $t$ can be denoted as $\mathbf{a}(t) =[\mathbf{w}(t),\mathbf{p}(t),\Psi(t),\mathbf{f^l}(t),\mathbf{f^e}(t)]$, where $\mathbf{w}(t) = [w_{10}(t),...,w_{NM}(t)]$ is bandwidth allocation vector, $\mathbf{p}(t) = [p_{1}(t),...,p_{N}(t)]$ is transmission power vector, $\mathbf{\Psi}(t) = [\Psi_{0}(t),...,\Psi_{M}(t)]$ is the vector associated with the   computation task that leaves edge servers, $\mathbf{f^l}(t) = [f_{1}^l(t),...,f_{N}^l(t)]$ and $\mathbf{f^e}(t) = [f_{10}^e(t),...,f_{NM}^e(t)]$ are the vector of computation resource that edge servers allocate to devices. Taking network stability constraint into account,  the optimization problem for  minimizing $\eta_{EE}$ can be formulated as:
\begin{subequations}
	\label{S1}
	\begin{align}
	\centering
	&P1:~~~\min_{\mathbf{a}(t) }	~~\eta_{EE}  \notag\\    	              
			s.t.~
	          &\sum_{i\in\mathcal{U}} \dfrac{w_{ij}(t)} {w_j}\leq1, ~~ w_{ij}(t)\geq 0 \label{c1}\\ 
			&~0\leqslant p_{i}(t) \leqslant p_{i,max}(t),\label{c2}\\    	
			&~0\leqslant f_{i}^l(t) \leqslant f_i^{l},\label{c3}\\ 	&~\sum_{i\in\mathcal{U}}f_{ij}^e(t)\leq f_j^e,~~ f_{ij}^e(t)\geq 0 \label{c4}\\
			&\Psi_j(t)*c\leq f_j^e\tau, ~~  \Psi_j(t)\geq 0 \label{c5}\\
			& (\ref{st1})-(\ref{st2}) \notag
	\end{align}
\end{subequations}

Constraint (\ref{c1})  is the bandwidth allocation constraint.  
Constraints (\ref{c2})  and (\ref{c3}) denote the transmission power and computation resource constraints, respectively. Constraint (\ref{c4}) ensures that the sum of the computation resource of each base station allocated to all devices does not exceed the total amount of computation resource the base station has. Constraint (\ref{c5}) implies that  the amount of computation resource for  processing task $\Psi_j$  cannot exceed the available computation resources. 
Constraints (\ref{st1}) and (\ref{st2})  guarantee that  the stability of all task queues. 

Since computation resource, transmission power and bandwidth need to be determined at each time slot, P1  is  a stochastic optimization problem, which is challenging to solve by applying classic convex optimization algorithms such as interior-point method and Lagrangian duality theory.  
 From Eq. (\ref{time_SBS}) and Eq. (\ref{energy_SBS}),  wireless communication rate and allocated  computation resource jointly determine the energy consumption of edge server execution. 
Thus, the radio resource management problem is coupled with the computation resource allocation problem. Moreover,  in radio resource management problem, bandwidth and transmission power are also highly  coupled. The complex coupling among optimization variables and mixed combinatorial feature makes it difficult to solve P1. Further, the stochastic task arrival, dynamic channel state information and dynamic task buffer make designing an efficient  resource management policy for devices and edge servers quite challenging.

Lyapunov optimization is a powerful methodology for solving optimization problems with long-term objective and constraints, which requires less prior information  on the task arrival, channel state information, and  task buffer.   The principal idea  behind  Lyapunov optimization is to  transform the optimization problem with  long-term objective into a series of  subproblems with short-term objective, and to transform the long-term constraints into constraints with queue stability.   Besides, Lyapunov optimization is
of low computational complexity by optimizing  subproblem through an online algorithm.    In this paper, we exploit Lyapunov optimization to transform the original stochastic optimization problem as a deterministic per-time block problem and propose a stochastic computation offloading algorithm to solve P1.

\subsection{Lyapunov-based Problem Transformation and Digital twin-predicted Perturbation}

To construct a Lyapunov optimization framework, we add a perturbation vector $ {\beta}  = [\beta_1,...,{\beta}_N]$ in Lyapunov function to keep the value of this function always small.  The perturbation parameters are simulated in the digital twin and it will be given in the following.
 We define the quadratic Lyapunov function as the sum of squared queue backlogs,
\begin{equation}
L(\Theta(t)) = \dfrac{1}{2}\{\sum_{i\in \mathcal{U}} [Q_i^l(t) -\beta_i]^2+\sum_{j\in \mathcal{B}}  Q_j^e(t)^2\}
\end{equation}
where $\Theta(t) = [Q^l(t),Q^e(t)]$ represents current  task queue lengths of devices and edge servers.  
 Further, we define the conditional  drift as
\begin{equation}
\bigtriangleup L(\Theta(t)) = \mathbb{E} [L(\Theta(t+1))-L(\Theta(t)) | \Theta(t)]
\end{equation}
By minimizing $\bigtriangleup L(\Theta(t))$, we can short the length of task  queues towards a smaller value. 

Accordingly, the Lyapunov drift-plus-penalty function can be expressed as
\begin{equation}
\bigtriangleup_VL(\Theta(t))=\bigtriangleup L(\Theta(t)) +V\mathbb{E}[\eta_{EE}(t)|\Theta(t)]
\end{equation}
where $\eta_{EE}(t) =  E^{tol}(t)/\sum_{i \in \mathcal{U}}\sum_{j \in \mathcal{B}}(D_i^t(t)+D_{ij}^e(t))$, $V$ is a non-negative weight parameter. By minimizing $\bigtriangleup_VL(\Theta(t))$, we can ensure network stability, and meanwhile minimize network EE. We derive the upper bound of    $\bigtriangleup_VL(\Theta(t))$ as, 
 \begin{equation}
 \label{cal0}
 \begin{split}
 &\bigtriangleup_V L(\Theta(t)) \leq C - \sum_{i\in\mathcal{U}} [Q_i^l(t)-\beta_i]\mathbb{E}[\Psi_i(t)-\lambda_i(t)|\Theta(t)]\\&-\sum_{j\in\mathcal{B}}Q_j^e(t)\mathbb{E}[\Psi_j(t)-\sum_{i\in\mathcal{U}}D_{ij}^e(t)|\Theta(t)] \} +V\mathbb{E}[\eta_{EE}(t)|\Theta(t)].
 \end{split}
 \end{equation}
 where $C = \dfrac{1}{2}\{\sum_{i\in\mathcal{U}}[\Psi_{i,max}^2 +\lambda_{i,max}^2]+\sum_{j\in\mathcal{B}}[\Psi_{j,max}^2+(\sum_{i\in\mathcal{U}}D_{ij,max}^e)^2 ] \}$ and $\Psi_{i,max}, \lambda_{i,max}, \Psi_{j,max}$ and $D_{ij,max}^e$ are the upper bounds of $\Psi_{i}(t), \lambda_{i}(t), \Psi_{j}(t)$  and $D_{ij}^e(t)$, respectively.

Based on the Lyapunov optimization theory, we can minimize the right side of the inequality in (\ref{cal0}) to obtain the optimal solution of P1. According to \cite{huang2012utility}, perturbation vector $\beta$ is very important as it influences the performance of optimization of the designed algorithm.  Based on the definition of perturbation vector in \cite{huang2012utility}, $\beta$ is the maximal lower bound of local task queue. Since digital twin is a mirror of physical network, it can easy get any information of the network and predicts each one of  $\beta$ based on
\begin{equation}
\label{pretur}
\beta_i = V \eta_{EE}'(t) + \Psi_{max}
\end{equation} 
where $\Psi_{max} = \max({\Psi_i(t)})$.

Thus, we first utilize digital twin to simulate the perturbation parameter of devices and then optimize the right side of the inequality in (\ref{cal0})  in each time slot.  The proposed stochastic computation offloading algorithm for DTN is shown in Algorithm \ref{DTN}, where P2 needs to be solved per-time slot. The traditional method to solve P2 is to decompose it into several sub-problems and  alternatively solving sub-problems until it converges to the global optimal solution. However, when wireless channels change or task queues update, each sub-problem needs to recalculate the optimal solution. Frequent operations to solve sub-problems will influence the convergence. DRL is an emerging technique which can find a near optimal solution in a real-time manner. {Thus, we  design a DRL-based algorithm to find the optimal solution of P2. }

 \begin{algorithm}[!tbp]
 	\caption{The Stochastic Computation Offloading Algorithm with Digital Twin-predicted Perturbation for DTN}
 		\label{DTN}
 	\begin{algorithmic}[1]
 \State  At the beginning of each time slot, digital twin first predicts perturbation vector $\beta$ based on local task queue and Eq. (\ref{pretur}), 
  \State   Digital twin observes $\Theta(t)$ and $\lambda_{i}(t)$ and determines $a(t)$ by solving the following problem in each time slot,
  \begin{equation}
   \small
  \begin{split}
  P2: &\min_{\mathbf{a}(t) }~~ V [E^{tol}(t)-\eta_{EE}(t)\sum_{i \in \mathcal{U}}\sum_{j \in \mathcal{B}}(D_i^t(t)+D_{ij}^e(t))]+\sum_{j\in\mathcal{B}}\\&\{Q_j^e(t)[\sum_{i\in\mathcal{U}}D_{ij}^e(t)-\Psi_j(t)] \} - \sum_{i\in\mathcal{U}} [Q_i^l(t)-\beta_i][\Psi_i(t)-\lambda_i(t)] \\&
   ~~~s.t. ~~~ (\ref{st1})-(\ref{st2}), (\ref{c1})-(\ref{c5})
  \end{split}
  \end{equation}
  \State Updates $Q_i^l(t)$ and $Q_j^e(t)$ based on Eq. (\ref{queue1}) and (\ref{queue2}), 
  \State $ t = t+1$.
 	\end{algorithmic}
 \end{algorithm}

\section{DRL-empowered Stochastic Computation Offloading Algorithm for DTN}
\begin{figure*}
 	\centering
 	\includegraphics[width =6.50 in]{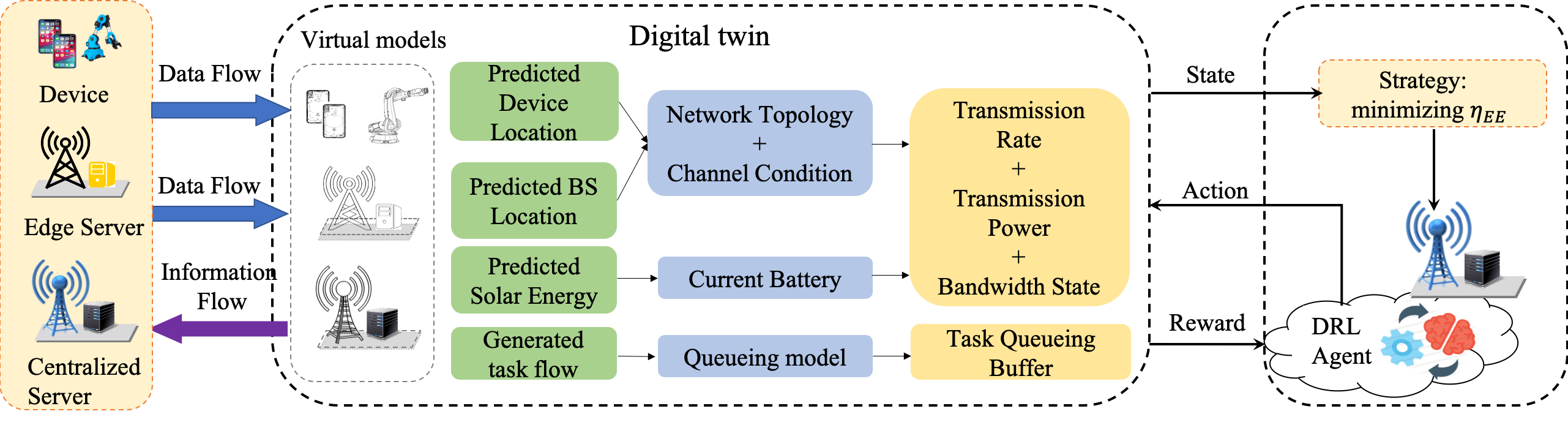}
 	\caption{Digital twin enabled DRL  algorithm}
 	\label{DT_framework}
 \end{figure*}

The framework of the digital twin enabled DRL  algorithm is illustrated in Fig. \ref{DT_framework}. Digital twin mirrors the network topology and parameters of physical wireless network and transmits network state to  DRL. DRL derives the best strategy to minimize network energy efficiency. 

\subsection{Digital Twin-simulated Network Environment}

To solve P2, the system first constructs Markov decision process, i.e., $\mathcal{M}=(\mathcal{S},\mathcal{A},\mathcal{P},\mathcal{R})$, and then use DRL algorithm to explore actions. 
From Fig. \ref{DT_framework}, the network state $s(t)$ is constructed by digital twin and outputted to DRL agent.

{To gather network environment information, digital twin needs to predict location, energy and the generated task flow of devices and base stations. Digital twin can adopt the existing K-Nearest Neighbors (KNN) classification  method and position prediction algorithm in \cite{li2019mobile} to predict users' location.} 
The maximal current transmission power $p_{i,max}(t)$ at time slot $t$ is the summarize of the predicted energy and the $p_{i,max}(t-1)$. The generated task flow is based on the application running on each device. 
After gathering the network information, digital twin updates  network topology, channel condition and queueing model. The main operation in the update of network topology is to make user association decision, which decides the connection between devices and base stations. Here we adopt online user association method in \cite{ao2017approximation}.  
Then, digital twin generates current state and transmits it to the DRL agent.  

Thus, at the beginning of time slot $t$, the DRL agent construct system state which includes  transmission data rates between devices and base stations, available bandwidth,  computation resources,  transmission power, and queue length. We can define system state $s(t)\in\mathcal{S}$ at time slot $t$  as 
\begin{equation}
s (t)=  \{\mathbf{R}(t), \mathbf{F}, \mathbf{p}_{max}(t),\mathbf{w},  \Theta(t)\}. 
\end{equation}
The state space $\mathcal{S}$ is as follows:
\begin{itemize}
 \item $\mathbf{R}(t) =\{[R_{11}^s(t),...,R_{1M}^s(t)],...,[R_{N1}^s(t),...,R_{NM}^s(t)],$ $[R_{10}^m(t),...,R_{N0}^m(t)]\}$: $N\times(M+1)$  wireless data rate matrix where  $R_{ij}^s(t) \geq 0$ and $R_{i0}^m(t) \geq 0$;
\item $ \mathbf{F}= [f_1^l,...,f_N^l ,f_0^e,..,f_M^e]$: $1\times(N+M+1)$ computation resource vector where $f_{i}^l \geq 0 $ and $f_{j}^e \geq 0 $;
\item $\mathbf{p}_{max}(t) = [p_{1,max}(t) ,...,p_{N,max}(t)]$: $N\times1$ transmission power  vector  at time slot $t$;
\item $\mathbf{w}= [w_0,...,w_j]$:  $1\times (M+1)$  bandwidth vector;
\item $\Theta(t)= [Q^l(t),Q^e(t)]$, where $Q^l(t) = [Q_1^l(t),...,Q_N^l(t)]$ indicates the queue length of the local task buffer and $Q^e(t) = [Q_0^e(t),...,Q_M^e(t)]$ is the queue length of the task buffer on edge servers.
\end{itemize}

 Since $\mathbf{a}(t) =[\mathbf{w}(t),\mathbf{p}(t),\Psi(t),\mathbf{f^l}(t),\mathbf{f^e}(t)]$ in P2 denotes system operation at time slot $t$,  we define it  as the  output from the DRL agent (i.e., action).   The action space $\mathcal{A}$ includes following fields:
\begin{itemize}
\item  $\mathbf{w}(t) = [w_{10}(t),...,w_{NM}(t)]$:  $N\times(M+1)$  bandwidth allocation matrix where $w_{ij}(t) \in[0, w_j]$;
\item $\mathbf{p}(t) = [p_{1}(t),...,p_{N}(t)]$: represents $N\times 1$  transmission power vector where $p_i(t) \in[0, p_{i,max}(t) ]$;

\item $\mathbf{\Psi}(t) = [\Psi_{0}(t),...,\Psi_{M}(t)]$:  $1\times (M+1)$  vector where $0 \leq \Psi_{j}(t)\leq f_j^e\tau/c$ is the computation task that leaves edge server $j$;
\item $\mathbf{f^l}(t) = [f_{1}^l(t),...,f_{N}^l(t)]$:  $N\times1$  computation resource allocation vector where $f_{i}^l(t) \in[0,f_i^l]$;
\item $\mathbf{f^e}(t) = [f_{10}^e(t),...,f_{NM}^e(t)]$ :   $N\times(M+1)$   computation resource allocation matrix where $f_{ij}^e(t) \in[0,f_e^j]$;
\end{itemize}
It is worth noting that all variables in action  $\mathbf{a}(t) $  are  continuous. Thus,  we will utilize a policy gradient-based DRL algorithm to explore policy.

After executing action $\mathbf{a}(t) $, digital twin updates system state and estimates  immediate reward $\mathcal{R}^{imm}(s(t),a(t))$.  In a traditional Markov decision process, the system updates its state based on the given  transition probability $Pr(s(t+1)|s(t),a(t))$. However, in DRL, the distribution of transition probability is often unknown. The DRL agent utilizes deep neural network to approximate it.

The immediate reward function is defined as the objective of P2 problem, i.e.,
\begin{equation}
\label{reward1}
\small
\begin{split}
&\mathcal{R}^{imm}(s(t),a(t))= - V [E^{tol}(t)-\eta_{EE}(t)\sum_{i \in \mathcal{U}}\sum_{j \in \mathcal{B}}(D_i^t(t)+D_{ij}^e(t))]\\&+\sum_{i\in\mathcal{U}}\{Q_j^e(t)[\Psi_j(t)-\sum_{i\in\mathcal{U}}D_{ij}^e(t)] \} + \sum_{i\in\mathcal{U}} [Q_i^l(t)-\beta_i][\Psi_i(t)-\lambda_i(t)] 
\end{split}
\end{equation}
 After computing  immediate reward, the system updates its state from $s(t)$ to $s(t+1)$ based on action $a(t)$. 
 
  The objective of the DRL agent is to maximize  the cumulative reward, 
 \begin{equation}
 \small
 \label{long}
\mathcal{R} = \max \mathbb{E}\left[\sum_{t=0}^{T-1} \delta^t\mathcal{R}^{imm}(s(t),a(t))\right],
 \end{equation}
 where $\delta\in[0,1]$ is the discount factor. 
 If all tasks  are   satisfying  the constraints of P2, DRL agent gets a total reward. Otherwise, the agent receives a penalty,  which is a negative  constant.

\subsection{Asynchronous Actor-Critic  Algorithm}
The DRL  algorithm is classified into value-based and policy gradient-based. Value-based DRL algorithms, such as DQN and double DQN,  estimate Q-values and $\epsilon$-greedy strategy to explore policy with
discrete action space. But  value-based DRL algorithms are of limited value for problems with continuous action space. Policy gradient-based DRL can learn stochastic policies effectively for tackling  problems with continuous action space problems. The main idea of this method is to optimize a parameterized stochastic policy by estimating the gradient of the expected reward of the policy and then updating the parameters of the policy in the gradient direction. We deploy AAC algorithm in digital twin to optimize cumulative reward $R$. 

AAC is an asynchronous learning algorithm which utilizes multiple agents to interact with its own environment and each agent contains a replica of the environment \cite{mnih2016asynchronous}.  A specific AAC agent is  \textit{Actor-Critic} based, where \textit{Actor} is used to generate actions and \textit{Critic} is used to evaluate and criticize the current policy by processing the reward obtained from the environment. 


1)  \textit{Actor-Critic based policy gradient training: } 


${a}(t) = \pi(s(t)|\theta_\pi)$ denotes the policy learned from current state, where   $\pi(s(t)|\theta_\pi)$ is the explored offloading and resource allocation policy produced by deep neural network of actor network. 
The network parameter of actor network is denoted as $\theta_{\pi}$ and it is trained through the policy gradient method \cite{mnih2016asynchronous}. The gradient of the expected cumulative discounted reward is calculated by,
  \begin{equation}
  \label{pg}
\bigtriangledown\theta_{_{\pi}} \mathbb{E}_{\pi} [\sum_{t=0}^{\infty}\delta^{t}R^{imm}(t)] = \mathbb{E}_{{\pi}}[\bigtriangledown_{\theta_{\pi}}\log \pi(s(t)|\theta_\pi)A_{\pi}(s,a)]
  \end{equation}
 where $A_{\pi}(s,a)$ is the difference between the expected cumulative discounted reward starting from state $s$ when agent chooses action $a$ and follows policy $\pi$. Here,  $A_{\pi}(s,a)$ is called the advantage function which indicates whether things get better or worse than expected.  $A_{\pi}(s,a)$ is calculated using,
    \begin{equation}
  A_{\pi}(s,a)=  \mathcal{R}^{imm}(s(t),a(t))+\delta v_{\theta_{v}}(s(t+1))-v_{\theta_{v}}(s(t)),
    \end{equation}
  The parameter  $\theta_{\pi}$  is updated  based on:
 \begin{equation}
\theta_{\pi} = \theta_{\pi}+\alpha_{\pi}\sum_{t} \bigtriangledown_{\theta_{\pi}}\log \pi(s(t)|\theta_\pi)A_{\pi}(s,a)],
 \end{equation}
 where $\alpha_{\pi}$ is the learning rate of the  actor network.  
We use  critic network to estimate the cumulative discounted reward of each state following the current actor network’s policy, which is also expressed as the value of each state, $v_{\theta_{v}}(s(t))$. $\theta_{v}$ is the network parameter of critic network.  The parameter $\theta_{v}$  is updated as follows:
 \begin{equation}
 \begin{split}
\theta_{v} = \theta_{v}+{\alpha_{v}}\sum_{t}\bigtriangledown_{\theta_{v}}&( \mathcal{R}^{imm}(s(t),a(t))+\\&\delta v_{\theta_{v}}(s(t+1))-v_{\theta_{v}}(s(t)))^2
 \end{split}
  \end{equation}
  where $\alpha_{v}$ is the learning rate of the critic network. 
 
 After the value function approximation $v_{\theta_{v}}(s(t))$ and parameter $\theta_{v}$ are updated by critic process, actor network then uses the advantage function $A_{\pi}(s,a)$ outputted from the critic process to update its policy parameters.
 
 2)  \textit{Asynchronous Learning with Experience Replay:} 

In  DQN  algorithm, there is an  important  component, i.e., replay memory, which disrupts the correlation between the experiences  such that  the sequence in DRL meets the independent and identical distribution. However, replay memory needs an off-policy learning algorithm to generate experiences and needs large amount of memory to store the generated experience.  AAC is an online DRL algorithm  that can reduce correlation between adjacent samples  by asynchronous learning with considerably less amount of computation. To implement AAC,  DTN sets up a global agent and multiple learning agents.The algorithms to be carried by learning agent and global agent are given in Algorithm \ref{iner} and Algorithm \ref{outer}, respectively.  Learning agent is deployed at SBS and it can interact with its own environment and the environment of all agents has same settings and structures. The learning agents parallelly explore and accumulate offloading and resource allocation policy.  After every $t_{max}$ learning steps, agents will send the accumulated updates  to the global  agent. The  global agent is deployed in MBS and it asynchronously updates $\theta_{\pi}$ and $\theta_{v}$.  $T_{max}$ represents the max training episodes. 
 \begin{algorithm}[!tbp]
 	\caption{Asynchronous  Actor-Critic  algorithm for Each Learning Agent}
 		\label{iner}
 	\begin{algorithmic}[1]
 	\State Assume  agent parameter $\theta'_{\pi}$ and  $\theta'_{v}$   
 	\State Initialize  learning step counter $t = 1$;
 	\Repeat
 	\State Synchronize agent parameter $\theta'_{\pi} = \theta_{\pi} $ and $\theta_{v}$   
 	\State Update global shared counter $T$
 	 \State $t_{start} = t$
 	 \State Use digital twin to construct current state $s(t)$
 	 	\Repeat
 	 	  \State Perform action $a(t)$ for problem P2 based on policy $\pi(s(t)|\theta'_{\pi})$
 	 	   	 	  \State Transfer to new state $s(t+1)$ and calculate immediate reward ${R}^{imm}(s(t),a(t))$
 	 	   	 	   \State $t \leftarrow t+1$
 	 	   	 	   \State $T \leftarrow T+1$
 	 	 \Until $t-t_{start}  == t_{max}$
 	 	 \State $R = v_{\theta'_{v}}(s(t))$
 	 	 \For {$i \in \{t-1,...,t_{start}\}$}
 	 	   \State $reward$.append(${R}^{imm}(s(t),a(t)) +\delta R$)
 	 	 \EndFor
 	 	 \State send $reward$ to global agent
    \Until $T > T_{max}$
 	\end{algorithmic}
 \end{algorithm}

 \begin{algorithm}[!tbp]
 	\caption{Asynchronous  Actor-Critic  algorithm for Global Agent}
 		\label{outer}
 	\begin{algorithmic}[1]
 \State  Assume global shared parameter $\theta_{\pi}$ and  $\theta_{v}$, and global shared counter $T'=0$
    	\While{ receive $reward$ from agent}
     	 	 \For {$i \in \{t-1,...,t_{start}\}$}
     	 	   \State Accumulate gradients with respect to $\theta'_{\pi}$ and  $\theta'_{v}$:
     	 	   \State $d\theta_\pi\leftarrow d\theta_\pi +\alpha_{\pi}\sum_{i} \bigtriangledown_{\theta'_{\pi}}\log \pi(s(t)|\theta'_\pi)A_{\pi}(s,a)]$
     	 	   \State $ d\theta_{v} \leftarrow d\theta_{v}+{\alpha_{v}}\sum_{i}\bigtriangledown_{\theta'_{v}} A_{\pi}^2(s,a)$
     	 	 \EndFor
     	 	 \State perform asynchronous update of $\theta_{\pi}$ using $d\theta_{\pi}$ and $\theta_v$ using $d\theta_{v}$ 
     	 	 \If {$T > T_{max}$}
     	 	  \State break
     	 	 \EndIf
        \EndWhile
 	\end{algorithmic}
 \end{algorithm}

\section{Numerical Results}
\label{sr}

{ We consider a network topology with one MBS,} $M=3$   SBSs, and $N=20$ devices. We consider Rayleigh fading channels.
The maximum transmission power of devices is set to $100$ mW.   The noise power is $\sigma^2 = 10^{-11}$ mW. The bandwidth of the MBS and each SBSs are $10$ MHz  and $5$ MHz, respectively.  In addition, $ \tau= 100$ ms, $c= 100$ cycles/bit.   The CPU computation capabilities of the devices, the SBSs and the MBS, are $ 0.5$, $10$, and $50$ GHz, respectively.  The actor network of AAC has three fully-connected hidden layers each with 128 neurons whose activation function is ReLU and an output layer with 8 neurons using softmax function as the activation function. The critic network has three fully-connected hidden layers each with 128 neurons whose activation function is ReLU and one linear neuron as output. We use Python and TensorFlow to evaluate the performance of our proposed stochastic computation offloading  algorithm.   { Based on the definition of immediate reward (\ref{reward1}),  the minimization of P2  is equivalent to the maximization of DRL reward.  For ease of observation, we define the objective of P2 as system cost and the system cost equals to the negative value of the DRL cumulative reward.}

  \begin{figure}
       	\centering
       	\includegraphics[width =3.4 in]{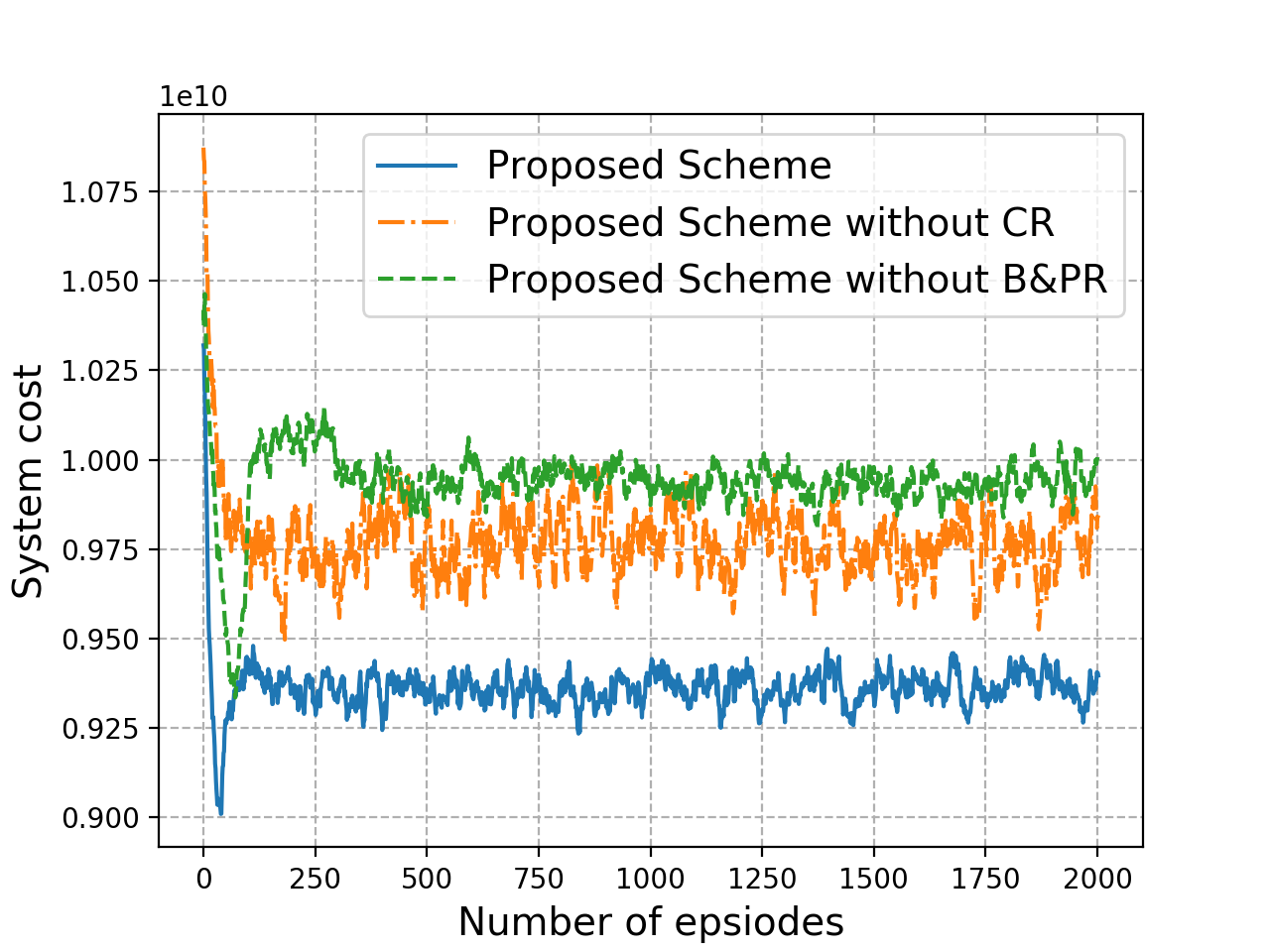}
       	\caption{System cost under different schemes.}
       	\label{fig1}
       	         \vspace{-0.1 in}
       \end{figure}

  Fig. \ref{fig1} illustrates the relationship between system cost and training episodes under different schemes.  The blue curve represents joint optimization of computation offloading,  bandwidth and transmission power, but without computation resource allocation.  The green
  curve shows the performance of the proposed scheme with computation offloading and computation resource allocation optimization  but without  bandwidth and transmission power allocation. From Fig.  \ref{fig1} we can see the performance of the red curve outperforms the two benchmarks, since it can concurrently optimize computation offloading,  bandwidth and transmission power, and computation resource allocation.   Besides, the system cost of the blue curve is lower than  the system cost of the green curve. This means,  compared with the optimization of computation resource, the optimization of bandwidth and transmission power has a greater influence on the performance.

      \begin{figure}
           	\centering
           	\includegraphics[width =3.4 in]{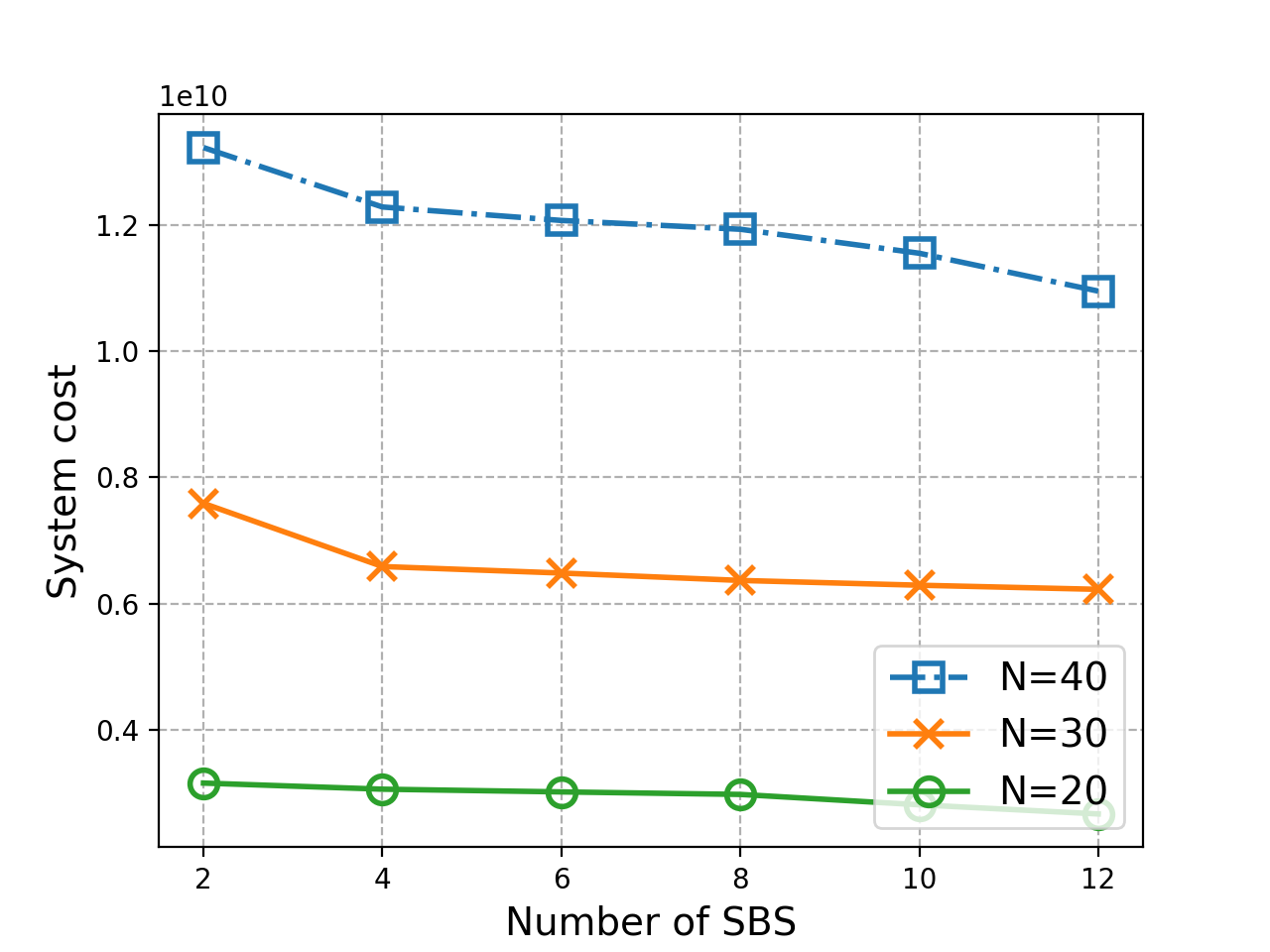}
           	\caption{ System cost with respect to the number of SBSs}
           	\label{fig2}
           	         \vspace{-0.1 in}
           \end{figure}

Fig. \ref{fig2} plots the comparison of  the system cost with respect to the number of SBSs. We observe that when $N=40$, the value of system cost  decreases with the increase of the number of SBS. When $N=20$,   the value of system cost  does not change much with the increase of the number of SBS.   This indicates, when the number of devices is large, increasing the number of SBSs can reduce system cost. When the number of devices is small, increasing the number of base stations has little effect on reducing system cost.

   \begin{figure}
           	\centering
           	\includegraphics[width =3.5 in]{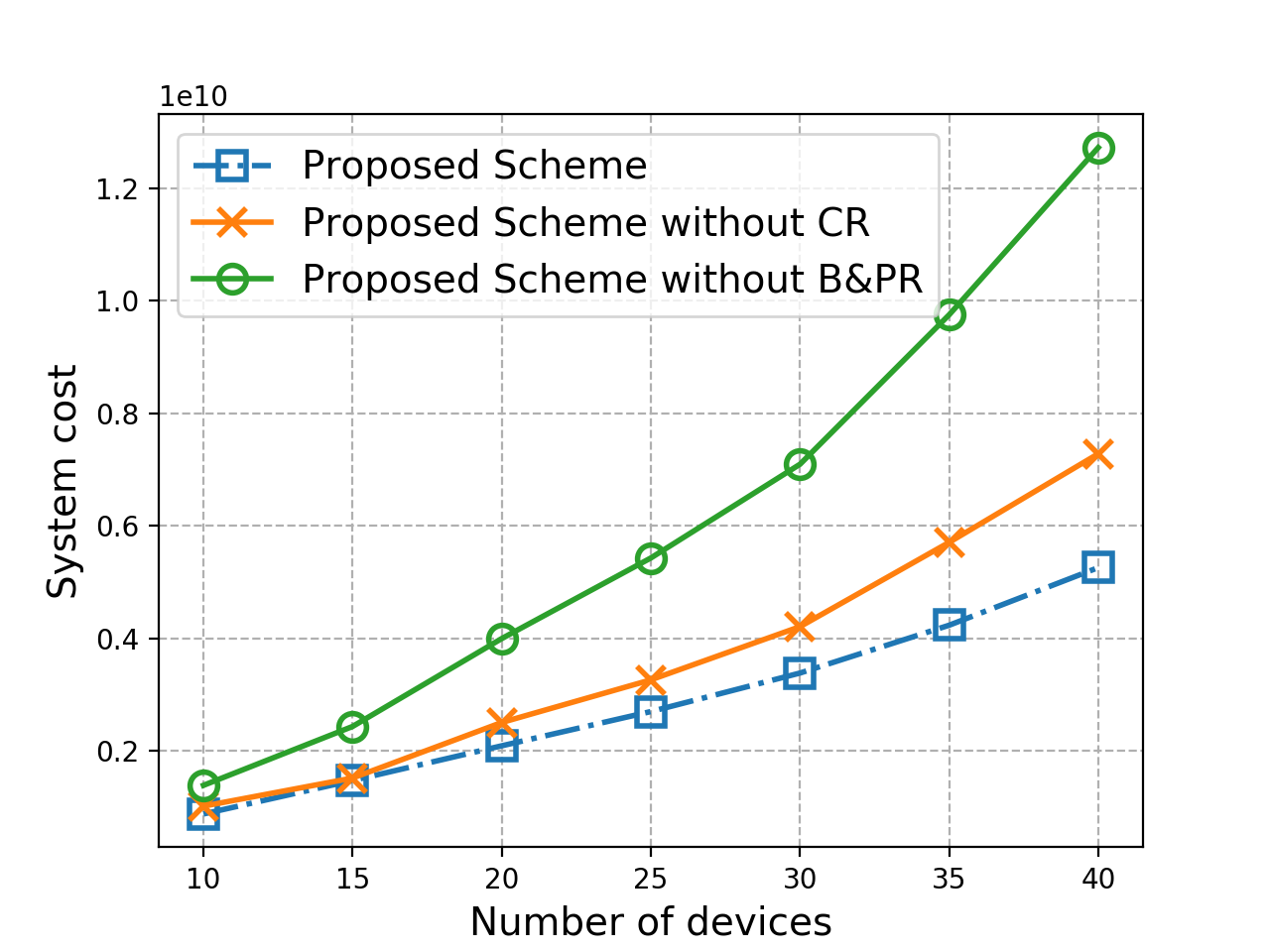}
           	\caption{System cost with respect to the  number of devices under different schemes.}
           	\label{fig7}
           	         \vspace{-0.1 in}
           \end{figure} 
    
 Fig. \ref{fig7} shows the comparison of  the system cost with respect to the number of devices under different schemes. The number of devices ranges from 10 to 40. From Fig. \ref{fig7},  we can draw several observations. First, the system cost of three  different schemes respectively increases as the number of devices becomes large.  The reason is that the growth of devices leads to more offloading requests, which results in the consumption of  more communication and computation resource. Second, the performance of the proposed algorithm outperforms two benchmarks by jointly optimizing computation offloading,  bandwidth and transmission power, and computation resource allocation.

    \begin{figure}
         	\centering
         	\includegraphics[width =3.5 in]{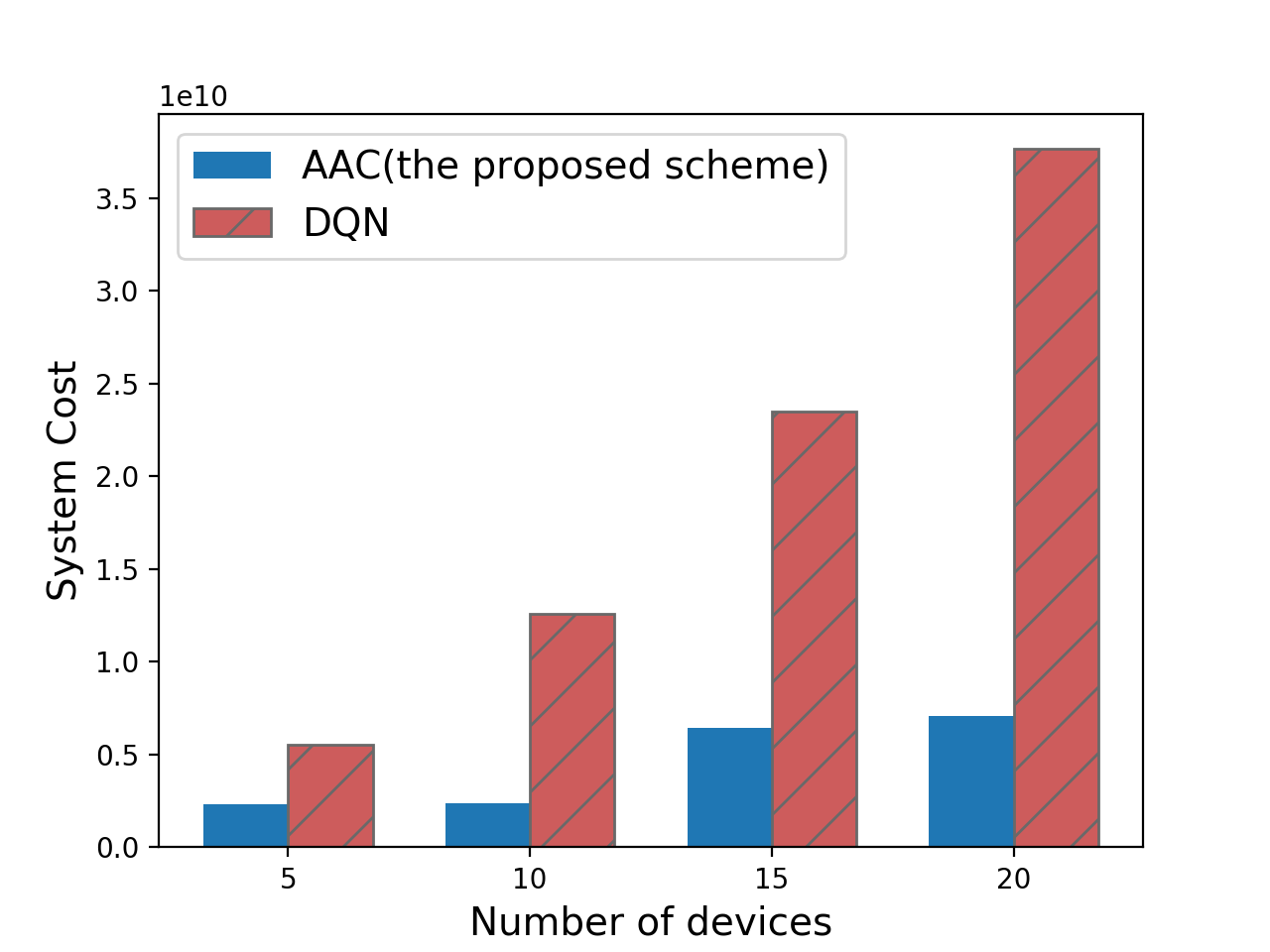}
         	\caption{System cost with respect to  the number of devices under different DRL algorithms.}
         	\label{fig3}
         	         \vspace{-0.1 in}
         \end{figure}  
              
          \begin{figure}
               	\centering
               	\includegraphics[width =3.5 in]{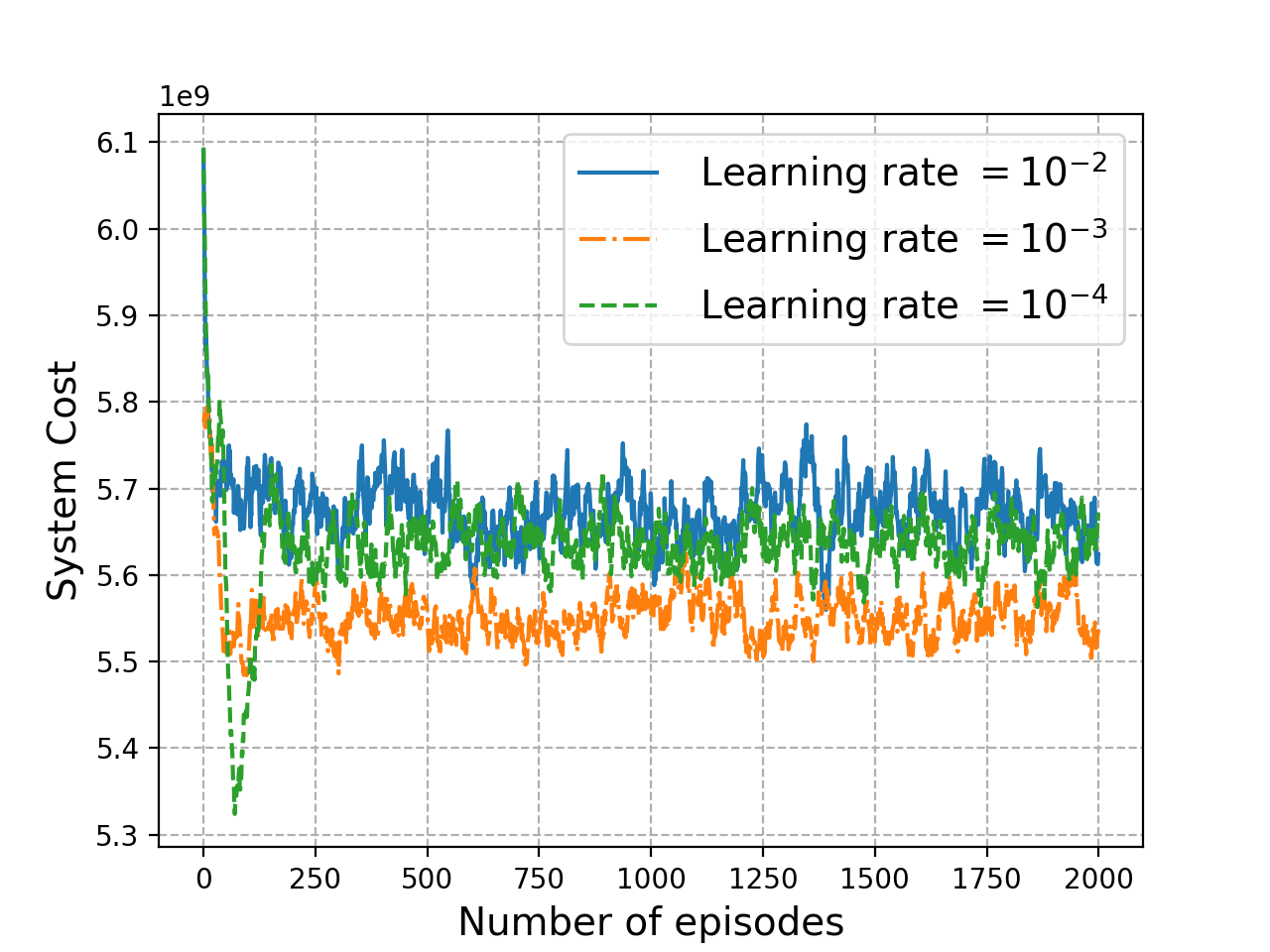}
               	\caption{Impact of the learning rate on performance.}
               	\label{fig4}
               	         \vspace{-0.1 in}
               \end{figure}

 Fig. \ref{fig3} illustrates the comparison of  system cost with respect to number of devices under different DRL algorithms. The number of devices ranges from 5 to 20. We can observe that the system cost increases with the increase in  the number of devices.  For given amount of computation  resources,   a large number of devices results in a low task execution rate and high energy  consumption. This inevitably  increases system cost. Moreover, our AAC based algorithm performs considerably better compared to the DQN. The main reason is that action discretization in DQN may lead to skipping  better actions.  Fig. \ref{fig4} shows the impact of learning rate on the performance of the proposed algorithm. We can see, when learning rate is $0.001$, the system cost of the proposed algorithm converges to the lowest value. Thus, $0.001$ is the best learning rate for  the proposed algorithm.
\section{Conclusions}
\label{c}
 In this paper, we proposed a DTN architecture for IIoT, which utilizes digital twin to construct the network topology and stochastic task arrival model in IIoT networks.	 Then, we formulated  the stochastic computation offloading and resource allocation problem to jointly optimize offloading decision, transmission power, bandwidth and computation resource.   As the formulated problem is a  non-convex stochastic programming problem, we  leverage the Lyapunov optimization technique to equivalently transform the original problem to  a deterministic per-time slot problem. Finally, we utilized AAC algorithm to  solve the computation offloading  and resource allocation problem. Numerical results demonstrate that our proposed  algorithm  significantly outperforms the benchmarks.
 
	\bibliographystyle{IEEEtran}
	\bibliography{reference}
	\end{document}